# Energy-efficient and Robust Cumulative Training with Net2Net Transformation

*Aosong Feng, and Priyadarshini Panda\**
*Department of Electrical Engineering,*
*Yale University, USA, 06510*
*\*priya.panda@yale.edu*

**Abstract**— *Deep learning has achieved state-of-the-art accuracies on several computer vision tasks. However, the computational and energy requirements associated with training such deep neural networks can be quite high. In this paper, we propose a cumulative training strategy with Net2Net transformation that achieves training computational efficiency without incurring large accuracy loss, in comparison to a model trained from scratch. We achieve this by first training a small network (with lesser parameters) on a small subset of the original dataset, and then gradually expanding the network using Net2Net transformation to train incrementally on larger subsets of the dataset. This incremental training strategy with Net2Net utilizes function-preserving transformations that transfers knowledge from each previous small network to the next larger network, thereby, reducing the overall training complexity. Our experiments demonstrate that compared with training from scratch, cumulative training yields ~2x reduction in computational complexity for training TinyImageNet using VGG19 at iso-accuracy. Besides training efficiency, a key advantage of our cumulative training strategy is that we can perform pruning during Net2Net expansion to obtain a final network with optimal configuration (~0.4x lower inference compute complexity) compared to conventional training from scratch. We also demonstrate that the final network obtained from cumulative training yields better generalization performance and noise robustness. Further, we show that mutual inference from all the networks created with cumulative Net2Net expansion enables improved adversarial input detection.*

## I. Introduction

Deep learning neural networks have emerged as a powerful tool in various fields to perceive, detect and classify different forms of data [1, 2]. On the one hand, larger datasets such as, CIFAR100 [3], ImageNet [4] have been collected to evaluate performance of neural networks; on the other hand, researchers have come up with deeper neural network architectures, such as VGG [5] and ResNet [6] to deal with such datasets. Generally, machine learning algorithms receive a fixed dataset as input, initialize a new neural network with no prior knowledge, and then train that model to convergence by repeated iterations of forward and backward propagation on the entire dataset. Thus, training a large network (with millions of parameters) from scratch using large datasets requires the full model to be stored and updated during each iteration of training. This consumes considerable storage, memory and computational resources.

A stream of work in building efficient networks is through knowledge distillation [7] that enables small low memory footprint networks to mimic the behavior of large complex networks. Net2Net technique proposed by Chen *et al.* [8] is an inverse variant of the knowledge distillation technique that serves as an efficient way to train a significantly larger neural network from a small neural network. Net2Net uses *function preserving transformations* to expand small neural networks into wider or deeper networks while preserving and transferring the knowledge from the previously trained small networks into each larger version. Net2Net accelerates the training process and also improves the performance of the network on large datasets.

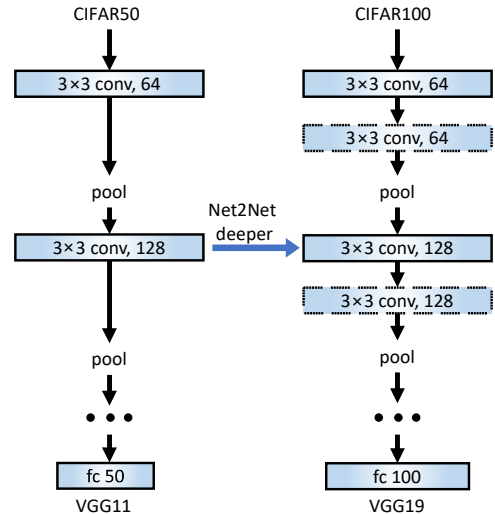

Fig. 1. The cumulative training process using Net2Net transformation. VGG11 is expanded to VGG19 with Net2Net deeper transformation while the number of neurons in the output layer increase from 50 to 100.

In this paper, we propose a cumulative training strategy with Net2Net transformation that achieves training computational efficiency without incurring large accuracy loss, in comparison to a *baseline* model trained from scratch. Instead of utilizing the full dataset and network with desired size throughout the training process, we first train a small network (with lesser parameters) on a small subset of the original dataset. Then, we gradually expand the small network towards the desired size using Net2Net transformation with incremental training on larger subsets of the original dataset, as shown in Fig.1. This cumulative training strategy with Net2Net utilizes *function-preserving transformations* that transfers knowledge from each previous small network (say, $Net_1$ or VGG11 in Fig. 1) and sub-dataset (say, $D_1$ or CIFAR50 in Fig. 1) to the next larger network (say, $Net_2$ or VGG19 in Fig. 1) and larger sub-dataset (say, $D_2$ or CIFAR100 in Fig. 1 where, $D_1 \in D_2$), thereby, reducing the overall training complexity. One of the key advantages of using Net2Net with cumulative training is that the new, larger network ($Net_2$) immediately performs as well as the previous network

($Net_1$) on the new sub-dataset ($D_2$), rather than spending time passing through a period of low performance. Our experiments show that the proposed cumulative training strategy yields better training efficiency with improved robustness and generalization performance (empirically evaluated with noise resilience and ablation study [9]). We also combine pruning with cumulative Net2Net training that yields a final neural network with compact configuration (thus, delivering reduced inference complexity) while being training-efficient. Please, note, all reductions and improvements with our proposed strategy is considered with respect to a baseline 'trained from scratch' model. Finally, we also find that ensemble or mutual inference that utilizes the output from all the sub-networks (obtained with cumulative training) to make the final prediction can be beneficial to detect and resist against adversarial inputs.

## II. RELATED WORK

### A. Incremental learning

Incremental learning, also known as life-long or evolutionary learning, refers to online learning strategies which work with limited memory resources [10]. They focus on how to learn in a streaming setting, in which case the network will continuously use the new input data to extend its knowledge when classes of new labeled data are available. For example, support vector machines have been used for incremental learning by training a new classifier for each new stream of data and finally combining all the classifiers to make the final decision [11]. Some methods also focus on learning new tasks from new samples using transfer learning technique [12, 13]. Li *et al.* [14] also proposed the concept of 'learning without forgetting'. They address the issue of using only new data to train the network while keeping its original capabilities. Our method is different from incremental training because we show both the original and new data during retraining and expansion (hence, termed as '*cumulative training*'), and, we focus on the energy efficiency aspect of training.

### B. Networks that share information

Sharing part of the neural network is a good way to do knowledge transfer between different models. For example. Pham *et al.* [15] use parameter sharing to improve the efficiency of network architecture search by forcing all the child models to share the weights. Luong *et al.* [16] use encoder-decoder sharing to combine multi-task learning with sequence to sequence learning for natural language processing. Partial network sharing strategy for 'learning without forgetting' [7] and tree-based efficient neural architectures [24, 25] have also been proposed. In such works, the authors keep part of the network intact and retrain remaining part only on new data. Our cumulative training scheme uses a similar philosophy. All the models generated during the iterative Net2Net expansion process share part of the information of their common classes, and we can transfer such information with function preserving transformations. Note, we show both the old and new data at the retraining stage, in order to improve the training convergence. A key difference between our methods and other tree or partial network sharing methods is that, we transform the same network end-to-end as we grow the network and the dataset. Other methods, generally add new branch of classifiers iteratively to the present network hierarchy to perform partial network sharing.

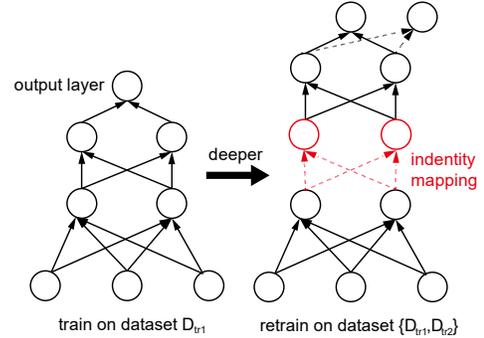

Fig. 2. Illustration of Net2Net expansion strategy. The deeper expansion will add new layer that are initialized as identical mapping on top of the original convolutional layer. Note, adding a convolutional layer will also incur a new batchnorm layer. The weights represented by dash line in the last output layer is randomly initialized.

In general, incremental and network sharing strategies mentioned above ultimately increase the training complexity of the model as they add additional data without appropriate knowledge transfer from previous configuration. Our cumulative approach enables knowledge transfer at zero accuracy loss due to Net2Net expansion and thus, gives a key benefit of lower training complexity and faster convergence.

### C. Net2Net training

Our scheme uses the Net2Net expansion strategy [8]. The main difference between [8] and our cumulative method is that we split the entire dataset, and introduce a new sub-dataset into the current training dataset every time we expand the model with Net2Net. We continue this till the final network is trained on the whole dataset. In contrast, Net2Net training [8] uses the full dataset throughout the expansion and training process. Our approach shows that the function preserving transformations also carry information from one data sub-data domain to another if trained in a cumulative manner.

## III. EXPERIMENTAL METHODOLOGY

### A. Net2Net methodology

Many network families have similar network architectures within a family. For example, VGG19 has similar block structure as VGG11, with 6 more convolution layers. Therefore, we can expand VGG11 to VGG19, and instead of adding the convolutional layers with random initialization, we use the Net2Net deeper technique (see. Fig. 2) that performs expansion while preserving the knowledge from trained VGG11. Such expansion eliminates the need to train the new VGG19 model from scratch again to attain the previous accuracy. The Net2Net deepening technique allows us to transform any net into a deeper one. It replaces a layer $h^{(i)} = \Phi(h^{(i-1)}W^{(i)})$ with two layers $h^{(a)} = \Phi(U^{(i)T}\Phi(W^{(i)T} h^{(i-1)}))$, as shown in Fig. 2, where $\Phi$ is the activation function and $W$ is the weight matrix for a given layer $i$. The new matrix U is initialized to an identity matrix, but remains free to learn to take any value later. For deepening convolutional layers, we set the convolutional kernels to be zero-surround filters with central value as 1 and remaining 0.

Note, while we use deepening in this work, we can also utilize Net2Net widening technique [8] that widens the layer size during the expansion process.

A key variation of our proposed cumulative training from that of standard Net2Net [8] is that the output layer also needs to be expanded with desired units. Our expansion strategy is accompanied with incremental training as we switch from one sub-dataset to another during Net2Net expansion. We find that we cannot simply use a wider Net2Net transformation (see [8] for details) on the softmax output layer as it interferes with the training convergence. Hence, for the final classifier layer, the newly added connections are randomly initialized (see Fig. 2) and are eventually learnt during the training process. As a result, our method expands the network as well as its capability from $K_1$ to $K_1 + K_2$ output classification (in Fig. 1, we go from a 50-class to 100-class classifier).

### B. Cumulative Learning with Net2Net

The algorithm of our training strategy is shown below.

---

**Algorithm** 1: Pseudo-code for Cumulative Training with Net2Net

---

**Input:** Training and Testing datasets $D_{tr}$, $D_{te}$ with target labels
**Output:** Final model learnt with cumulative Net2Net expansion
1. Split $D_{tr}$ and $D_{te}$ into sub-datasets $\{D_{tr1}, D_{tr2}, \ldots, D_{trN}\}$ and $\{D_{te1}, D_{te2}, \ldots, D_{teN}\}$ where $D_{tr,te_i} \in D_{tr,te_{i+1}}$
2. Initialize the base network $Net_1$
3. Train $Net_1$ using $D_{tr1}$
4. **for** i = 2:N **do**
5.     $Net_i = Net2Netexpand(Net_{i-1})$
// Note, Net2Netexpand( ) can either be widening or deepening
6.     Expand the output layer of $Net_{i-1}$
// Output neurons added based on number of classes in $D_{tri}$
7.     Train $Net_i$ using $D_{tri}$
8. **end for**

---

We divide the whole training process into multiple stages. In the first stage, we train the small base network $Net_1$ using small dataset $D_{tr1}$. In the subsequent stages, we iteratively expand the trained neural network to the desired size using Net2Net scheme. Simultaneously, we increase the number of final output layer's neurons in order to accommodate the total number of classes in the new sub-dataset. Hence, we train a transformed network (say $Net_i$) using larger dataset (say $D_{tri}$ where, $D_{tri} = \{D_{tr1}, D_{tr2} \ldots D_{tri}\}$). We repeat this process across multiple stages of expansion until we finish training the whole dataset with the desired network shape. For example, in Fig. 1, instead of training VGG19 network on CIFAR100 from scratch, we first train a smaller VGG11 on CIFAR50, then expand VGG11 to VGG19 using Net2Net deepening, and use CIFAR100 to train it.

A noteworthy point here is that the sub-networks obtained with our technique are trained on the partial dataset in an incremental manner, where, the current dataset contains all previously shown data along with new data. It is the final network that will receive the full dataset as input. The aim of using Net2Net expansion with cumulative training is to effectively transfer the partial data knowledge from one network to another without loss in accuracy.

### IV. EXPERIMENTAL RESULTS

In this section, we present the results that demonstrate the energy efficiency and robustness of the cumulative training scheme. We conduct a series of experiments, primarily using CIFAR10, 100 and TinyImageNet dataset [18] on VGG networks of different depths. We imported github models and used similar hyperparameters and training methodologies as [19, 20] to conduct our experiments in PyTorch. In all our experiments, the baseline model refers to the model trained from scratch on the full dataset. We compare the efficiency and robustness of the final network obtained with cumulative Net2Net training to that of the corresponding baseline. Note, the final network obtained through Net2Net will be equivalent in size to the baseline and receive the full dataset as input.

### A. Training Efficiency

Fig. 3 compares the test accuracy vs. training epochs trend of cumulative Net2Net expansion scheme with that of baseline for CIFAR100 and TinyImageNet. For CIFAR, we expand the network as VGG11→VGG16→ VGG19 while incrementally expanding the dataset as CIFAR50→ CIFAR70→ CIFAR100. We see that there is minimal loss in accuracy (<1%) as we expand the network from one data domain (say, CIFAR50) to another (say, CIFAR70). This implies that cumulative training transfers information without loss of knowledge that eventually causes faster convergence at lower complexity. Importantly, Net2Net approach gives the same level of final accuracy as the baseline model trained from scratch with almost the same number of total training epochs. Note, the baseline model is a large VGG19 network that will incur the same fixed compute operations (OPS) throughout the training period. In contrast, our approach initially trains smaller networks on smaller datasets (that will incur lower OPS) and then progressively expands to larger datasets/networks. The OPS incurred with cumulative training is thus dynamic during the training process which eventually results in lower overall training complexity.

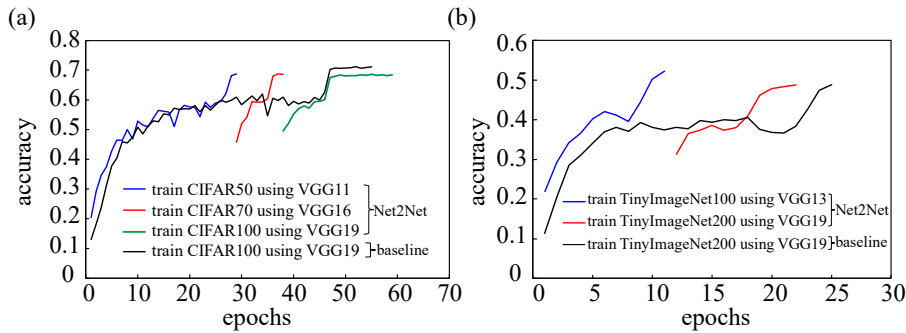

Fig. 3. Comparison between the cumulative training approach and baseline. The cumulative training scheme always converges faster towards the same accuracy level with lesser epochs for (a) CIFAR100 and (b)TinyImageNet datasets. The plots shown are for 70, 30 epochs of training.

Table 1: Summary of Results. The final accuracy obtained with cumulative training is **highlighted** in each scenario. The total training complexity (*M*) calculated from Eqn. 1 denotes the total compute incurred in T-Tera, P-Peta OPS. We show the accuracy, M for each sub-network ($Net_i$) trained on the sub-dataset as well as total M ($\Sigma_i M_{Net_i}$) compared to Baseline.

| Data | Cumulative Net2Net Training | | | | Baseline | |
|---|---|---|---|---|---|---|
| | Networks ($Net_i$) | Accuracy (%) | $M_{Net_i}$ | $M = \Sigma_i M_{Net_i}$ | Accuracy (%) | M (1x) |
| CIFAR5-10 | VGG11-16 | 87.4-**90.7** | 99.3T-313.0T | 412.3T (0.77x) | 90.7 | 532.1T |
| CIFAR50-100 | VGG11-19 | 68.7-**70.5** | 107.0T-383.0T | 490.0T (0.53x) | 70.2 | 919.2T |
| CIFAR50-70-100 | VGG11-16-19 | 68.7-68.7-**68.4** | 107.0T-98.6T-229.8T | 435.4T (0.47x) | 70.2 | 919.2T |
| CIFAR20-70-100 | VGG11-16-19 | 70.3-69.6-**70.7** | 23.0T-252.0T-421.3T | 696.2T(0.76x) | 70.2 | 919.2T |
| CIFAR20-50-70-100 | VGG11-13-16-19 | 70.3-69.7-70.1-**69.6** | 23.0T-79.8T-108.55T-287.3T | 499.6T(0.54x) | 70.2 | 919.2T |
| TinyImageNet100-200 | VGG13-19 | 52.4-**47.9** | 0.5P-1.4P | 1.9P (0.50x) | 47 | 3.8P |
| TinyImageNet 50-100-150-200 | VGG11-13-16-19 | 51.1-49.6-47.2-**46.4** | 0.2P-0.45P-0.75P-1.27P | 2.7P (0.71x) | 47 | 3.8P |

We define the training computational complexity *M* for cumulative training approach as

$$M = \Sigma_{i=1}^{N} \#Training\ Iterations_{Net_i} \times \#MAC_{Net_i} \quad (1)$$

Total *#MAC* quantifies the total number of Multiply and Accumulate operations (that translates to energy consumption) incurred in a given sub-network $Net_i$. For a sub-network $Net_i$, *#Training Iterations* specify the total number of training rounds required to reach saturating accuracy. In case of cumulative training, the total training complexity *M* is obtained by summing the individual complexities of training each sub-network. Note, the efficiency and robustness comparisons between our approach and the baseline are done for iso-accuracy. Essentially, we train both the baseline and each of the sub-networks obtained with Net2Net until training convergence or accuracy saturation occurs. Thus, the number of training iterations in both baseline and our Net2Net model might vary.

Our results are summarized in Table 1. Across different datasets, we obtain 0.5x-0.8x reduction in training complexity at near iso-accuracy (+/-0.5% difference) compared to the baseline. It is worth noting that the training complexity in our approach dynamically changes and each sub-network expends considerably lesser effort than the baseline. This can be attributed to the apt information transfer with Net2Net from one data domain to another which in some ways also improves the convergence behavior and overall accuracy.

We also observe that the complexity *M* is dependent on the number of intermediate expansion stages (say, *N*). For instance, CIFAR20-50-70-100 (*N=4, M=0.54x*) incurs higher complexity than CIFAR50-70-100 (*N=3, M=0.47x*). The general trend is that complexity *M* increases with *N*. Further, the starting point of the cumulative training procedure is also crucial to the end accuracy and *M*. For instance, having a smaller dataset in the beginning, such as, CIFAR20-70-100 scenario yields slightly better accuracy (70.7%) at higher *M* (0.76x) than CIFAR50-70-100 (accuracy = 68.4%, *M* = 0.47x).

*B. Pruning and Zero Mask Training*

Besides training efficiency, a key advantage of our cumulative training strategy is that we can perform pruning during Net2Net expansion to obtain a final network with compressed size (thus, lower inference compute OPS) compared to baseline. Here, we use the L1 norm based kernel pruning scheme and prune full filter kernels in the convolutional layers of a network [21]. Note, we do not prune the fully connected layers in this case. Essentially, during the cumulative training process, before applying the Net2Net expansion, we prune the filters of the learnt network from the current stage. We define a pruning ratio *R* which measures the ratio of total number of pruned kernels to the total number of kernels in the convolutional layers of the network.

Table 2 shows the parameter reduction, accuracy, total training complexity (*M*) for pruning with cumulative Net2Net training and the total inference OPS (quantified as #MACs) for different scenarios. We have a case corresponding to pruning while *keeping the Net2Net zero mask* intact. While expanding the convolutional layers with Net2Net technique, the filters are essentially identically mapped with a zero-surround kernel. For a 3x3 kernel, the central position is identity or 1 and the remaining 8 values are 0. Thus, the Net2Net expansion scheme inherently introduces a sparse zero mask kernel. The values of this new kernel can eventually take non-zero values during the learning process. While performing the pruning experiments, after pruning a network (*R=50%*) and then expanding it, we chose to fix the zero values in the expanded kernels during the next stage training. Note, in the zero mask case, we calculate the #MACs considering that our hardware contains a zero-checker logic that can discount the zero operations. As a result, #MAC in the pruning with zero mask case in Table 2 is lower than that of simple pruning.

Table 2: Summary of Pruning and Zero Mask Training. All (values) shown in brackets are comparisons with respect to a baseline model trained from scratch without pruning. For #parameters, M, #MAC, values <1 denote improvement.

| | Type | #Parameters | Accuracy (%) | M | # MAC |
|---|---|---|---|---|---|
| **CIFAR 50-100 VGG11-19** | Net2Net (R=0) | 20M (1x) | 70.5 (+0.3%) | 490.0T (0.53x) | 398.1M (1x) |
| | Net2Net (R= 50% for VGG11) | 15.41M (0.77x) | 70.4 (+0.2%) | 388.5T (0.42x) | 350.2M (0.88x) |
| | Keep Net2Net zero mask (R= 50% for VGG11) | 9.22M (0.46x) | 70.0 (-0.2%) | 334.9T (0.36x) | 152.9M (0.38x) |
| **Tiny ImageNet 100-200 VGG13-19** | Net2Net (R=0) | 20M (1x) | 47.9 (+0.9%) | 1.9P (0.5x) | 1.59G (1x) |
| | Net2Net ((R= 50% for VGG13) | 15.32M (0.77x) | 47.9 (+0.9%) | 2.04P (0.54x) | 1.4G (0.88x) |
| | Keep Net2Net zero mask (R= 50% for VGG13) | 10.61M (0.53x) | 46.5 (-0.5%) | 9.06P (2.37x) | 611.7M (0.37x) |

Table 3: Comparison of adversarial accuracy of models across different scenarios. The Net2Net expansion here follows Table 1.

| ε | CIFAR50-70-100 | | | TinyImageNet50-100-150-200 | | |
|---|---|---|---|---|---|---|
| | Net2Net (w/ MI) | Net2Net (w/o MI) | Baseline | Net2Net (w/ MI) | Net2Net (w/o MI) | Baseline |
| 0 | 69.83 | 68.40 | 70.20 | 47.38 | 46.43 | 47.43 |
| 0.005 | 60.86 | 59.66 | 57.99 | 37.53 | 33.96 | 35.13 |
| 0.01 | 53.79 | 50.65 | 47.98 | 30.21 | 24.91 | 25.63 |
| 0.02 | 40.58 | 36.77 | 34.71 | 20.69 | 13.5 | 13.72 |
| 0.05 | 22.90 | 17.76 | 12.93 | 9.00 | 3.07 | 3.44 |
| 0.1 | 13.18 | 9.03 | 8.20 | 3.56 | 0.66 | 1.59 |

For CIFAR50-100 case (corresponding to VGG11-19 expansion), simple pruning (with *R=50%*) and combined pruning with zero mask training results in good accuracy with improved *M* and overall inference OPS. In TinyImageNet100-200 case, simple pruning (*R=50%*) does not affect the accuracy/complexity. However, we observe a drastic increase in *M* to reach the same level of accuracy as the baseline or Net2Net (R=0, R=50%) when we apply zero mask intact condition. This means that as the complexity of data increases, all the new kernels and weights added through the expansion procedure need to undergo learning to achieve good and fast training convergence.

### C. Robustness of cumulative training

We also conducted noise analysis and ablation study [9] to analyze the robustness of our cumulative training technique. Both noise and ablation resiliency have been shown to characterize the overall generalization capability of a network [9]. Noise analysis was done by adding gaussian noise to the input images and monitoring the accuracy of the network with increasing noise variance. Fig. 5(a) compares the noise resiliency of a model trained with cumulative training (VGG11-19 on CIFAR50-100) against a baseline VGG19 trained on CIFAR100 fully. Fig. 5 (c) shows the resiliency of the networks (corresponding to CIFAR50-100 as above) when we ablate or zero out a fraction of units from the convolutional layers. In both cases we find that the baseline model has a sharper drop in accuracy than the model trained with Net2Net expansion. This means that our method yields more resilient networks. Similar robustness results were obtained for TinyImageNet100-200 scenario with VGG13-19 expansion (Fig. 5(b, d)).

### D. Mutual Inference for Adversarial Input and Error Detection

Another characteristic of Net2Net cumulative training strategy is that it will generate several different types of sub-networks with partial knowledge of the full dataset during training. Typically, we discard all the sub-networks and only use the final network obtained at the end of training for inference. Say, we have enough memory (such as in cloud servers) and can save the models generated, we can leverage all the sub-network's knowledge together with the final network to further improve the overall performance using an ensemble or mutual inference (MI) scheme. In cumulative training of CIFAR50-70-100, we obtain a final VGG19 trained using CIFAR100, a sub-network VGG11, VGG16 trained on CIFAR50, CIFAR70, respectively. For MI, we combine the predictions from the three networks using a weighted majority vote. Previously, with single model inference, we obtained 68.40% accuracy for VGG11-16-19 expansion on CIFAR50-70-100 and 46.43% accuracy on VGG11-13-16-19 expansion on TinyImageNet50-100-150-200. The MI scheme increases the accuracy of CIFAR model to 69.83% and TinyImageNet to 47.38%. It is evident that MI incurs higher inference OPS for a marginal improvement in accuracy. Then, the question arises if there any benefit to using MI.

The ability to detect adversarial samples is a very important capability to safeguard neural networks. We find that MI improves the adversarial robustness of networks. Table 3 compares the adversarial accuracy of Net2Net models with and without MI when exposed to adversarial test inputs created using FGSM attack [22] with varying attack strength ($\epsilon$). Net2Net without MI corresponds to the case when we use the final network obtained with cumulative training to make the prediction. All attacks are conducted assuming attacker has full knowledge of the model in the baseline and Net2Net without MI case. For Net2Net with MI case, we assume the attacker has no knowledge of the initial sub-networks but has complete knowledge about the final network. As shown in Table 3,

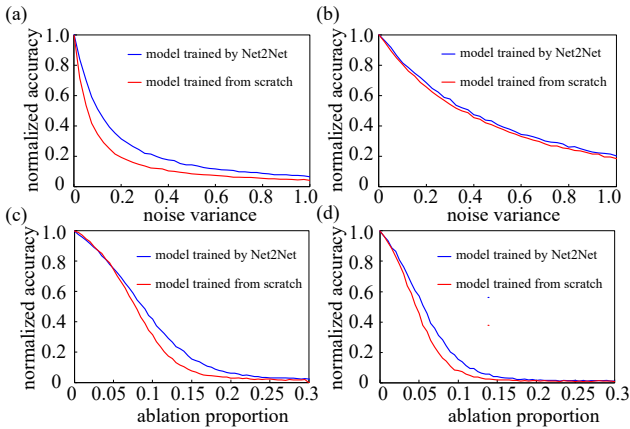

Fig. 4. Robustness of model trained with cumulative Net2Net training in comparison to baseline is shown. (a), (c) show the results of noise variation and ablation study for VGG11-19 CIFAR50-100 scenario, respectively. (b), (d) show the corresponding results on VGG13-19 TinyImageNet100-200.

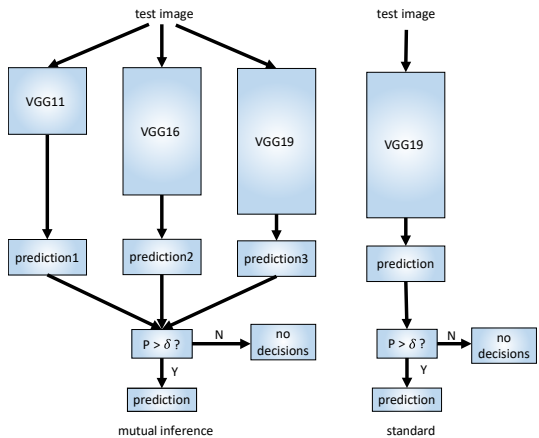

Fig. 5. Here, VGG19 is expanded from VGG16 and VGG11. In mutual inference, we use the information from all 3 networks to get the final result. To produce no-decision, generally, in standard case (final VGG19 obtained with Net2Net or baseline model), we compare the model's softmax confidence with a user-defined threshold $\delta$ to make a final decision. In mutual inference, we threshold the softmax output across all models that provides a stronger notion of no-decision.

Net2Net with MI yields higher accuracy than the other cases as $\epsilon$ increases.

We further use the MI scheme to detect adversarial inputs by enhancing the capability of the network to predict *no-decisions*. MI scheme for *no-decision* prediction is illustrated in Fig. 5. We use a thresholding technique similar to that of Hendrycks *et al.* [23] to test adversarial input detection ability. The inputs which are either correctly classified or classified as no-decision contribute towards good decisions. The inputs which are misclassified are considered as bad decisions. Therefore, the input prediction with thresholding as shown in Fig. 5 for both standard and MI case can fall into three buckets: (a) Inputs which are correctly classified (b) Inputs which are classified as 'no-decisions' (c) Inputs which are incorrectly classified. We report False Negative Rate (FNR) and True Negative Rate (TNR) to evaluate the adversarial detection ability of the Net2Net model with and without MI. The no-decision prediction in Net2Net without MI is conducted using standard model inference shown in Fig. 5. Our goal is to increase TNR while keeping FNR as low as possible. By using different threshold, we get the plot of TNR vs. FNR in Fig. 6. For a given FNR, TNR of Net2Net with MI is much higher than the TNR of Net2Net without MI. We believe that the partial knowledge and sharing of information between the sub-networks with MI improves the no-decision prediction capability. Note, in comparison to baseline, a model obtained with Net2Net cumulative training (both with or without MI) has higher TNR/FNR ratio.

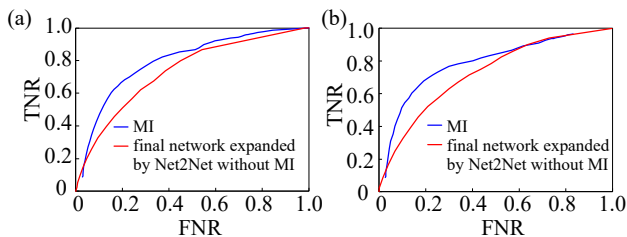

Fig. 6. By changing the threshold, we can get the relationship between TNR and FNR for (a) CIFAR50-70-100 and (b) TinyImageNet50-100-150-200 with and without mutual inference. The curve which approaches the left upper corner of the plot yields better adversarial detection ability.

## V. CONCLUSION

In this work, we propose a cumulative training scheme using Net2Net which incrementally expands the network and the dataset at the same time. Compared with training a model on the entire dataset from scratch (or baseline), our cumulative training method yields better training efficiency. This efficiency gain is due to the fact that our approach trains smaller networks on partial datasets and gradually increases the network size and the dataset while performing good knowledge transfer without any loss in performance. Combining Net2Net expansion with pruning, we show that cumulative training can be leveraged to obtain a high performing final network with compressed configuration and therefore less inference complexity. The generalization performance and robustness to noise of the final model obtained by cumulative training was also demonstrated to be better than the baseline model. Finally, we show that we can use all the models generated during the cumulative training expansion process for mutual or ensemble inference. Mutual inference enhances the robustness to adversarial attack and improves error detection ability with no-decision prediction.

## REFERENCES

[1] Y. Netzer et al., "Reading digits in natural images with unsupervised feature learning," In *NIPS*, 2011.
[2] G. Hinton et al., "Deep neural networks for acoustic modeling in speech recognition," *Signal Processing Magazine*, 2012.
[3] A. Krizhevsky, "Learning multiple layers of features from tiny images," Tech Report, 2009.
[4] J. Deng et al., "ImageNet: A large-scale hierarchical image database," In *CVPR*, 2009.
[5] K. Simonyan et al., "Very deep convolutional networks for large-scale image recognition," in *ICLR*, 2015.
[6] K. He et al., "Deep residual learning for image recognition," *arXiv preprint* arXiv:1512.03385, 2015.
[7] G. Hinton, et al., "Distilling the knowledge in a neural network," *arXiv:1503.02531*.
[8] T. Chen, et al., "Net2net: Accelerating learning via knowledge transfer," In *ICLR*, 2016.
[9] A. S. Morcos, et.al, "On the importance of single directions for generalization," in *ICLR*, 2018.
[10] A. Gepperth et. al, "Incremental learning algorithms and applications," in *ESANN*, 2016
[11] Y. M. Wen et. al, "Incremental learning of support vector machines by classifier combining," in *PAKDD,* 2007
[12] L. Fei-Fei, et al., "One-shot learning of object categories," *PAMI*, 2006.
[13] C. H. Lampert, et al., "Learning to detect unseen object classes by between-class attribute transfer," in *CVPR*, 2009.
[14] Z. Li et al., "Learning without forgetting," in *ECCV,* 2016.
[15] H. Pham, et al., "Efficient Neural Architecture Search via Parameter Sharing," in *ICML*, 2018
[16] M.-T. Luong, et al., "Multi-task Sequence to Sequence Learning," in *ICLR*, 2016.
[17] S. S. Sarwar et al., "Incremental learning in deep convolutional neural networks using partial network sharing," arXiv:1712.02719, 2017.
[18] Lucas Hansen, "Tiny imagenet challenge submission," *CS 231N*, 2015.
[19] [Online] https://github.com/kuangliu/pytorch-cifar
[20] [Online] https://github.com/tjmoon0104/Tiny-ImageNet-Classifier
[21] S. Han et al. "Learning both weights and connections for efficient neural network." In *NIPS,* 2015.
[22] G. Ian, et al., "Explaining & harnessing adversarial examples," in *ICLR*, 2015.
[23] D. Hendrycks et al., "A baseline for detecting misclassified and out-of-distribution examples in neural networks," in *ICLR*, 2017.

[24] P. Panda, et al., "FALCON: Feature driven selective classification for energy-efficient image recognition," *IEEE TCAD,* 2017.

[25] D. Roy, et al., "Tree-CNN: A hierarchical deep convolutional neural network for incremental learning," *arXiv:1802.05800*, 2018.